\documentclass{article}

\usepackage{PRIMEarxiv}
\usepackage{rotating}
\usepackage{lscape}
\usepackage{float}
\usepackage{multirow}
\usepackage[utf8]{inputenc} 
\usepackage[T1]{fontenc}    
\usepackage{hyperref}       
\usepackage{url}            
\usepackage{booktabs}       
\usepackage{amsfonts}       
\usepackage{nicefrac}       
\usepackage{microtype}      
\usepackage{lipsum}
\usepackage{fancyhdr}       
\usepackage{graphicx}       
\graphicspath{{media/}}     

\title{Towards a Taxonomy for the Use of Synthetic Data in Advanced Analytics
}

\author{
  Peter Kowalczyk, Giacomo Welsch, Frédéric Thiesse \\
  Chair of Information Systems Engineering \\
  University of Würzburg \\
  Würzburg\\
  \texttt{\{peter.kowalczyk, giacomo.welsch, frederic.thiesse\}@uni-wuerzburg.de} \\
}

\begin{document}
\maketitle

\begin{abstract}
The proliferation of deep learning techniques led to a wide range of advanced analytics applications in important business areas such as predictive maintenance or product recommendation. However, as the effectiveness of advanced analytics naturally depends on the availability of sufficient data, an organization's ability to exploit the benefits might be restricted by limited data or likewise data access. These challenges could force organizations to spend substantial amounts of money on data, accept constrained analytics capacities, or even turn into a showstopper for analytics projects. Against this backdrop, recent advances in deep learning to generate synthetic data may help to overcome these barriers. Despite its great potential, however, synthetic data are rarely employed. Therefore, we present a taxonomy highlighting the various facets of deploying synthetic data for advanced analytics systems. Furthermore, we identify typical application scenarios for synthetic data to assess the current state of adoption and thereby unveil missed opportunities to pave the way for further research.
\end{abstract}


\section{Introduction}
In the last decade, advanced approaches to the analysis and exploitation of large amounts of heterogeneous data (“big data”) have gained tremendous attention, particularly on the part of corporate decision-makers but also from academic researchers \cite{Delen2018, Holsapple2014, Mortenson2015}. The term “advanced analytics” generally refers to various methods beyond traditional multivariate statistics, mainly from the field of machine learning (ML), that leverage big data to drive decisions and actions (e.g., in organizations) \cite{Barton2012, Bose2009, Delen2018,Franks2013}. While researchers started to emphasize the suitability of these approaches mostly for (i) the design of innovative artifacts (e.g., decision support or process automation systems) and (ii) the induction of knowledge from quantitative studies \cite{Agarwal2014, Delen2018, Muller2016, Shmueli2011}, companies increasingly deploy analytics applications in order to exploit their promising business potential \cite{Barton2012, Franks2013}. Several research articles show that such applications—especially those driven by modern ML algorithms—may considerably improve efficiency and/or effectiveness in important business areas, such as predictive maintenance, financial fraud detection, capacity planning, and product recommendation \cite{Bose2001,Brynjolfsson2017,Chen2012,Goes2014,Grover2018,Jordan2015,Lecun2015,Minelli2013}. The average return on investment of modern data analytics applications in a business context is estimated at an almost inconceivable rate of 1,301\% \cite{Derstine2019}.

However, as the effectiveness of advanced analytics approaches naturally depends on the availability of the data to be analyzed, their application is highly challenging or even impossible in situations where data is limited or unavailable \cite{Berger2014,Nugroho2019}. Limited data availability is frequently observed across various use cases and relevant data sources in both research and practice. For instance, in a situation where historical consumer data are to be analyzed (e.g., for explaining consumer behavior or to generate appropriate product recommendations), the quality of the corresponding analysis heavily depends on the sheer amount of the data (i.e., the number of consumers, the length of the consumer history, and the features considered). Yet, from the perspective of a startup or a medium-sized enterprise, such massive datasets are usually not available. As a consequence, these organizations are inevitably at a disadvantage compared to larger competitors—especially those within the technology sector—who collect large datasets from diverse sources. In some cases, the former may have the option to remedy such disadvantages by acquiring the missing data within a legal scope. On the downside, this is typically associated with high costs. However, there are other circumstances in which the acquisition of relevant data is not possible, for example, if the data is subject to privacy restrictions (e.g., patients’ data in healthcare) \cite{Abouelmehdi2018,Iyengar2018,Patil2014}. Moreover, even in cases where the source of the data of interest is not external (e.g., if internal company data is to be analyzed), data might be limited due to a poor information technology infrastructure, faltering data streams, or the novelty degree of the area under investigation, among others \cite{Johnk2021}. In summary, if an organization’s ability to make use of advanced analytics is restricted by limited or unavailable data, the organization must choose between three options: either (i) spend substantial amounts of money on data, (ii) accept the shortcomings that are accompanied by analyzing limited data (if available), or (iii) entirely refrain from making use of advanced analytics techniques and thus dismiss the opportunity of exploiting the corresponding added value.

Against the backdrop of these adverse options, the proliferation of deep learning (DL) recently led to novel approaches that may help to overcome the barriers to the effective use of advanced analytics by being capable of generating artificial data algorithmically that can subsequently be analyzed. Such data is commonly referred to as synthetic data \cite{Nikolenko2019}. According to Kaloskampis et al. (2020) \cite{Kaloskampis2020}, El Emam et al. (2020) \cite{ElEmam2020}, and Gartner Inc. (2020) \cite{GartnerInc.2020}, synthetic data is created by generative models and intended to be highly realistic. In this way, synthetic data can be utilized to (i) generate entire samples, (ii) increase sample sizes, or (iii) anonymize data \cite{Abay2019,Nikolenko2019,Patki2016}. Moreover, some approaches to producing synthetic data even provide the option to include not yet encountered conditions into the data sample generated \cite{Bonnery2019}. Thus, synthetic data is a promising means to facilitate the application of advanced analytics, especially in situations where data is sparse. 

Despite the great potential of synthetic data, its practical use in driving advanced analytics is still very limited. We attribute this issue particularly to the novelty of the conceptual foundations of synthetic data, their application domains, as well as the variety of available data generation methods. To address this issue, we develop a taxonomy that is intended to support practitioners and researchers who are interested in using synthetic data in an advanced analytics scenario in (i) structuring and (ii) communicating their endeavor, as well as in (iii) selecting appropriate data generation techniques. For this purpose, we build upon the findings of recent studies that have already applied synthetic data in the context of various different research areas (e.g., computer science, physics, and biology). In the remainder of the paper, we (i) evaluate these articles by means of a structured literature review, which (ii) provides the foundation for our taxonomy. Subsequently, we (iii) derive several typical application scenarios for the use of synthetic data in advanced analytics—these reflect the current state of application of synthetic data and unveil several opportunities for future research.

\section{Background}
Generating synthetic data is no mold-breaking innovation \textit{per se} \cite{Gelman1992,Pomerleau1989}, but with the advent of DL and thus the various potentials of synthetic data application and generation, the topic regains importance \cite{Damer2018, Goodfellow2016, Kingma2019, Sharma2020, Wei2021}. Therefore, it is beneficial to develop an understanding of both the existing and emerging approaches driven by DL. To this end, this section provides a brief overview of these synthetic data generation approaches along with their respective advantages and disadvantages.

In essence, synthetic samples can either be derived (i) from prior knowledge (e.g., regarding statistical distribution parameters or in the form of pre-existing models and simulation tools) or (ii) by using some source data that is already at hand \cite{ElEmam2020}. While the former is highly dependent on the expertise involved to create adequate data, the latter is strongly affected by the quality of the data provided and how it is handled \cite{Wagner2021}. Here, the classical phrase \textit{garbage in}, \textit{garbage out} applies. More specifically, if no previous data cleansing and preparation has been conducted, the synthetic data generation method cannot provide valuable data. This usually pertains if essential aspects within the data (e.g., informative outliers) are missing (Wagner, 2021) \cite{Wagner2021}. Thus, output control is mandatory to ensure appropriate data generation. Moreover, the analyst has to carefully consider whether the effort required to employ a synthetic data generation method exceeds the effort of gathering empirical data (e.g., through surveys). The most commonly used data generation methods can be grouped into four categories, namely (i) statistical distributions, (ii) simulation models, (iii) data augmentation, and (iv) DL-based approaches \cite{Wagner2021}. In the following, we outline the four categories of methods for generating synthetic data.

\subsection{Statistical Distributions}
By observing the data from or by applying prior knowledge about the domain under consideration, it is possible to fit a statistical distribution model. Such models resemble or at least aim to approximate the actual probability distribution for the respective application field to create samples accordingly. The most commonly used statistical distributions include Gaussians \cite{Boor1999}, Chi-square \cite{Curran2002}, Lognormal \cite{Crow1988}, Student’s t \cite{Hofert2013,Kinderman1977}, Exponential \cite{Gray1994,Reiter2002}, or Uniform \cite{Gray1994}. Typically, only a few statistical parameters are necessary to fit a statistical distribution model (e.g., mean and standard deviation for a Gaussian). Therefore, a great benefit of this type of method is its quick and easy application. In addition, the parameters that correspond to a statistical distribution model are usually controllable, that is, they can be adjusted in order to meet the requirements at hand. Furthermore, the computational power required for sampling is typically smaller compared with the other sampling methods. On the downside, statistical distributions have disadvantages when it comes to modeling more complex dependencies and requirements into the data.

\subsection{Simulation Models}
In contrast to statistical distributions, simulation models rely on domain knowledge only for the configuration of a pre-defined environment to execute an activity-based simulation process and thereby generate observations that serve as data for later analysis. Consequently, this method allows to model and capture rather complex causal interactions between the variables in the system under consideration. Model-based approaches for creating synthetic data particularly include graphical simulators (e.g., by using game engines to generate traffic scenarios \cite{Cortes2020}) and agent-based exploration of pre-configured dynamics of a system \cite{Davidsson2001,Drogoul2003,Forrester1993,Forrester1994,Macal2010,Raberto2001}. Moreover, simulation models allow for incorporating time-based, observable states or immersing multiple agents into a simulation scenario in parallel \cite{Hare2004,Macal2010a}. Additionally, if such a simulation model is well configured once in advance for a specific task, it can easily be deployed over and over again. Apart from that, developing and sufficiently testing a new simulation model can hence be a challenging and resource-intensive task \cite{Midgley2007}. Without rigorous testing, undesired (or even worse undetected) interactions can occur, which may hamper efficient data sampling or lead to errors \cite{Isermann1999}.

\subsection{Data Augmentation}
Data augmentation refers to a class of transformation techniques that can be applied to source data to increase sample size or variety \cite{Shorten2019, Um2017}. Amidst the rise of DL, data augmentation gained much attention due to its simplicity and effectiveness. For example, to overcome limited data availability for bioimage analysis with DL, Segebarth et al. (2020) \cite{Segebarth2020} effectively deploy data augmentation as proposed by Falk et al. (2019) \cite{Falk2019}. Such transformations can be performed on various data types like images, text, time-series data, spectrograms, and tabular data. Images can be transformed via rotation, cropping, warping \cite{Beier1992}, jittering \cite{Hussain2017}, kernel filter based manipulation \cite{Davis1975, He2013, Prewitt1970}, color style mixing, random erasing and mixing of images (i.e., applying new backgrounds at random) \cite{Zhong2017}. Similarly, as for text, one can insert synonyms based on a dictionary and randomly swap or delete words \cite{Wei2020}. In terms of time-series data, common techniques are window-slicing (i.e., cutting sections), window or time-warping (i.e., slow down or speed up sections) \cite{Fawaz2018,Guennec2016}, jittering, or the rotation of time windows \cite{Rashid2019}. Next, spectrograms like audio data can be shifted vertically (i.e., pitch) or horizontally (i.e., time) \cite{Schluter2015}. Additionally, frequency filters can be applied or the intensity (i.e., volume in the case of audio data) can be tuned \cite{Schluter2015}. Tabular data can be augmented via oversampling techniques like the well-received synthetic minority oversampling technique (SMOTE) as introduced by \cite{Chawla2002}. Here, the goal is to encounter imbalanced data by directing the value sampling specifically to the underrepresented data classes \cite{Chawla2002,Fernandez2018}. For example, to systematically predict financial distress (i.e., external economic environment or internal financial decision failure) Sun et al. (2020) \cite{Sun2020} handle the rarity of such events by applying SMOTE. Aside from SMOTE multiple further developments in this regard emerged (e.g., SMOTEBoost \cite{Chawla2003}, ADASYN \cite{He2008}). A prerequisite for data augmentation—regardless of the respective data type under consideration—is the availability of some source data, that is, data augmentation is not possible without an initial dataset. Another challenging aspect associated with the use of pre-existing data is biases toward certain elements which might be naturally embedded into the data itself. Without adequate supervision, these biases could be reinforced via data augmentation leading to severe analytical errors \cite{Mehrabi2021}. 

\subsection{DL-based Approaches}
Aside from using DL to drive advanced analytics applications (e.g., image classification), some of the emerging DL approaches can in turn be used to generate synthetic data necessary for such analyses. As opposed to data augmentation, these DL-based approaches are capable of creating fully realistic, entirely new samples \cite{Chen2014,Fink2020,Lecun2015,Najafabadi2015,Park2015,Wang2020}. However, analogous to modeling with DL, such DL approaches to synthetic data generation rely on adequate data (i.e., in terms of sufficiency and quality) to create useful samples. Still, once trained sufficiently, a deep generative model is capable of providing as much data as requested. In addition, matured models can be easily shared or even repurposed through adaptation. DL-based approaches particularly include generative adversarial networks (GANs), variational autoencoders (VAEs), normalizing flows, energy-based models, hidden Markov models, Bayesian networks, and, finally, Boltzmann machines. GANs and VAEs in particular attracted increased attention recently for a wide range of application fields \cite{Damer2018,Goodfellow2016,Kingma2019,Sharma2020,Wei2021}. Therefore, we elaborate on these two methods in the following.

\begin{figure}[H]
  \centering
  \includegraphics[width=0.50\textwidth]{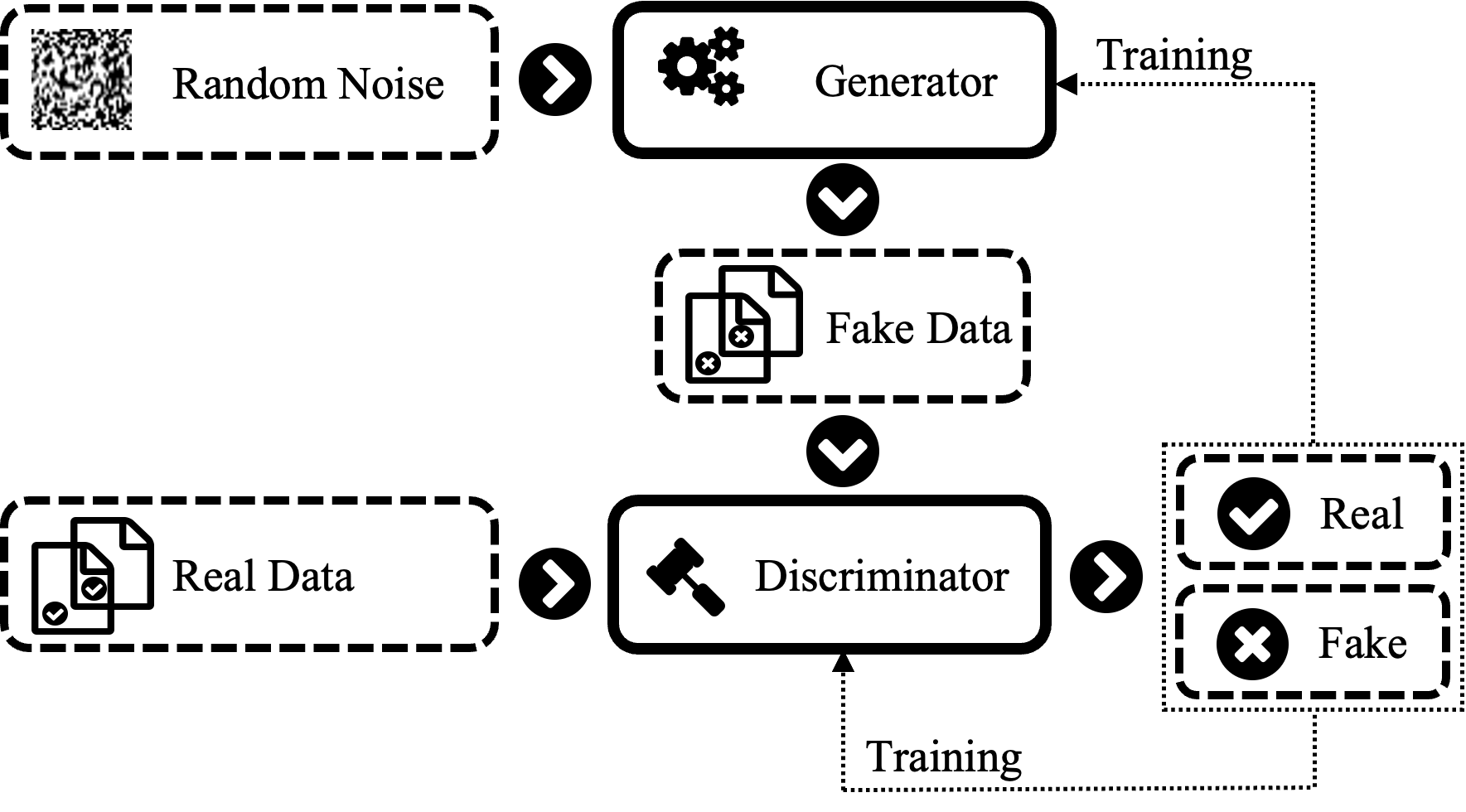}
  \caption{Overview on Generative Adversarial Network (based on Goodfellow et al. (2014) \cite{Goodfellow2014}).}
  \label{fig:fig1}
\end{figure}

Pre-trained GANs―as introduced by Goodfellow et al. (2014) \cite{Goodfellow2014}―are capable of processing inputs in random, unstructured form (i.e., Gaussian noise) to then apply a transformation such that entirely new samples are created. This is achieved by training two deep neural networks (i.e., the discriminator and the generator) contesting in the form of a zero-sum game (cf. Figure \ref{fig:fig1}).
 
While the discriminator classifies some input (i.e., either an original or a synthetic sample from the generator) as real or fake, the generator uses these classifications to generate more and more realistic data instances by trying to outsmart the discriminator by approximating the real joint distribution. The two opposing networks are trained over multiple epochs until ideally, the accuracy of the discriminator converges to 50\%, which means that real and fake samples are not discriminable anymore (i.e., the generator manages to provide realistic samples). GANs are considered to perform well on unstructured data and thus do not require the effort of previous data labeling. However, the parameters of GANs may underly heavy oscillations, which makes fine-tuning a model challenging \cite{Saxena2020}. Furthermore, the so-called “mode collapse phenomenon” (i.e., the generators outputs gradually become less diverse due to overoptimization for particular discriminator feedback) poses a severe difficulty when training GANs \cite{Saxena2020}. Still, because the concept behind GANs is rather new to academia, the research field is in constant flux, and new ideas to circumvent these issues are introduced repeatedly for a variety of domains \cite{Chen2021, PavanKumar2021}.

\begin{figure}[H]
  \centering
  \includegraphics[width=0.6\textwidth]{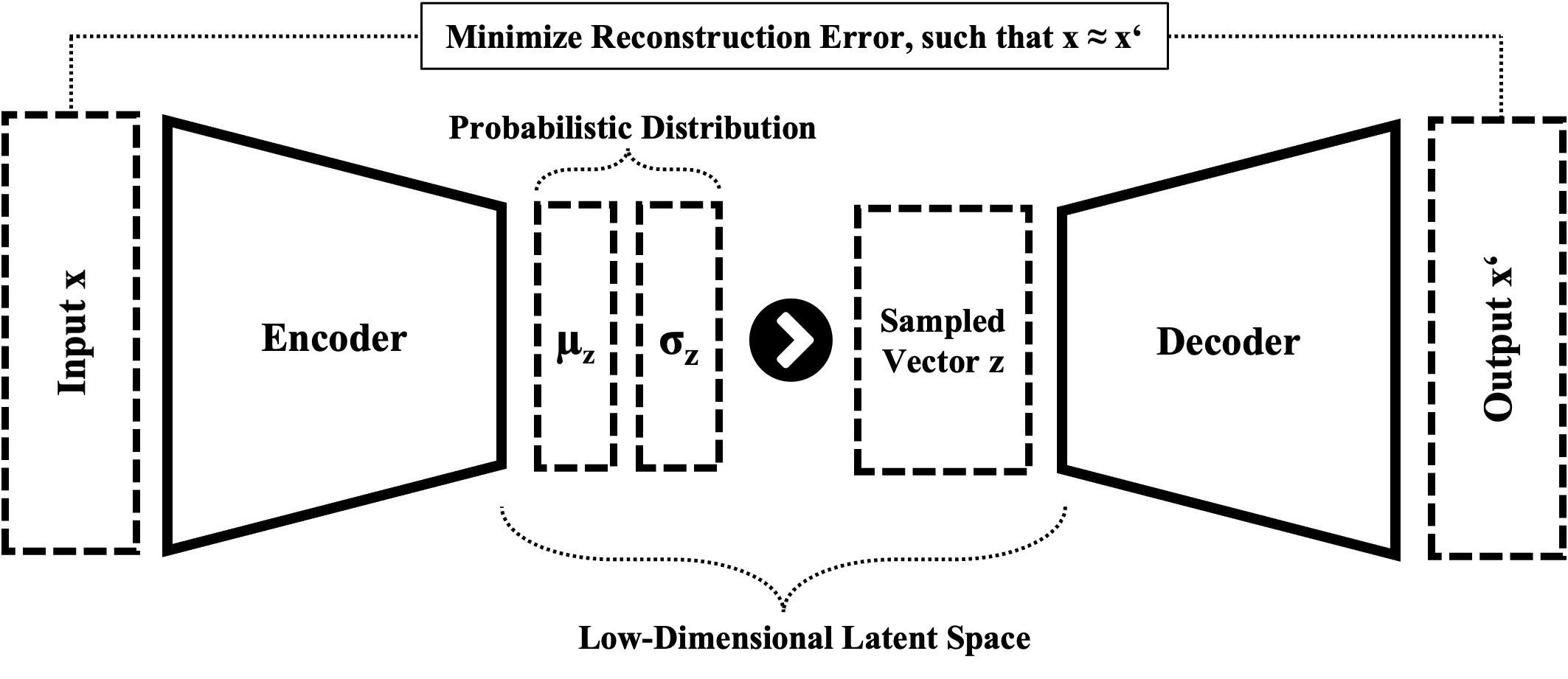}
  \caption{Overview of Variational Autoencoder (based on Kingma and Welling \cite{Kingma2014}).}
  \label{fig:fig2}
\end{figure}

VAEs are another DL-based approach to generating authentic synthetic data \cite{Kingma2014}. Again, two neural networks (i.e., one for encoding and the other for decoding) are coupled with each other, but unlike in GANs, these are not opposing each other but rather building upon each other in the form of a linear pipeline (cf. Figure \ref{fig:fig2}). At the outset, the encoder network converts source data into a dense latent representation space in the form of a distribution (i.e., mean and standard deviation vectors). Next, the decoder network is trained to reconstruct the data from this sparse latent distribution. Here, the reconstruction error (or generative loss), which indicates the difference between input x and output x’ can be computed. By introducing variance into the latent space, the training is regularized to prevent overfitting and ensure sufficient generative properties within the latent space. The degree of regularization is measured and guaranteed via the Kullback-Leibler divergence (i.e., the difference between the latent distribution and a standard Gaussian), which can be regarded as the latent loss. Both networks are optimized such that the combination of the quantifiable generative and latent losses is minimized, and the desired transformation function is obtained. Given such a transformation function, one can introduce random noise into the latent space to decode it and thus generate new samples \cite{Wan2018}. VAEs are suitable for processing a vast majority of data types, regardless of their characteristics―i.e., whether sequential or non-sequential, continuous or discrete, or finally labeled or unlabeled \cite{Simidjievski2019, Yu2019}. However, VAEs may learn uninformative latent representations \cite{Kim2018, VanDenOord2017, Yacoby2020} or unrealistic data distributions \cite{Tomczak2018,Yacoby2020}. 

\section{Research Approach}
\begin{figure}[H]
  \centering
  \includegraphics[width=0.85\textwidth]{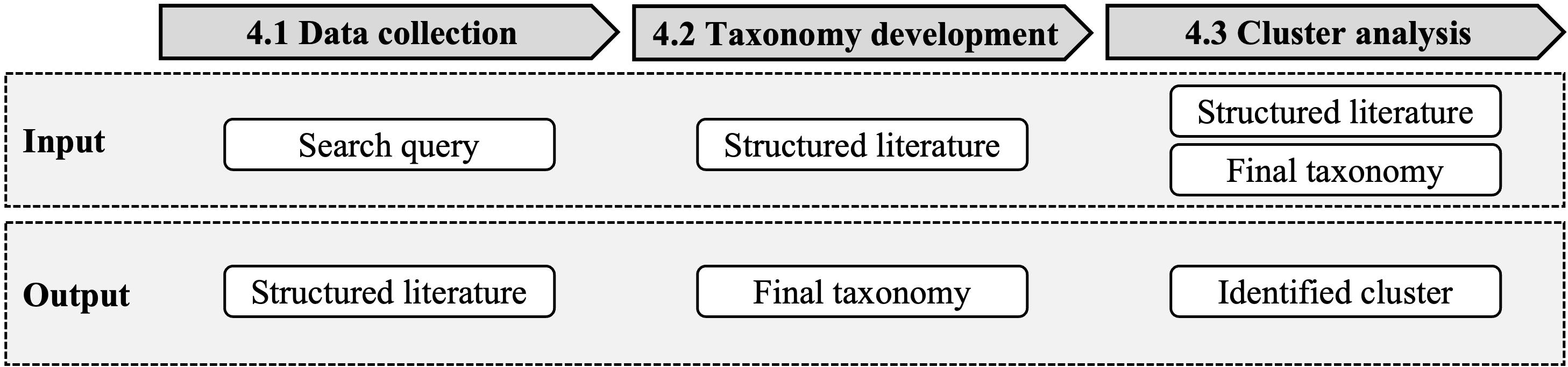}
  \caption{Research Approach in a Nutshell.}
  \label{fig:fig3}
\end{figure}

Against the background of the increasing importance of data-driven applications and the promise of synthetic data―especially in terms of the emerging DL-based generation approaches―we follow a three-step procedure (cf. Figure \ref{fig:fig3}) to develop a taxonomy regarding the use of synthetic data in different types of application scenarios and uncover the status quo.

At the outset, we collect literature that comprises studies which employ advanced analytics in combination with synthetic data. For this purpose, we conduct a structured literature review on the foundation of the well-received guidelines put forward by Vom Brocke et al. (2009) \cite{VomBrocke2009}. The guidelines refer to the following five tasks: (i) setting an adequate review scope, (ii) conceptualizing the topic, (iii) conducting the literature search, (iv) synthesizing the literature, and finally, (v) formulating a research agenda. However, as the literature is merely intended to serve taxonomy development, we refrain from synthesizing the literature exhaustively and from formulating a research agenda. Consequently, for the first task, in line with Vom Brocke et al. (2009) \cite{VomBrocke2009}, we make use of the taxonomy of Cooper (1988) \cite{Cooper1988} to determine the review scope. Next, we conceptualize the topic by specifying its key components based on the scope. In the consecutive task, we specify keywords that constitute the search query, which serves to search the different databases. In the fourth task, we systematically filter the articles retrieved according to their relevance by first removing duplicates and analyzing titles and abstracts as well as full texts. The remaining articles assist in developing the taxonomy. 

Next, we draw upon the well-established methodology for taxonomy development as presented by Nickerson et al. (2013) \cite{Nickerson2013}. According to the authors, a taxonomy assists researchers and practitioners in the systematical organization of knowledge for a specific domain and provides useful guidance for the exploration of relationships within emerging research fields \cite{Nickerson2013}. A taxonomy consists of a set of dimensions, each with multiple mutually exclusive characteristics, that is, a corresponding observation (e.g., a study) may only be assignable to one characteristic within any dimension \cite{Nickerson2013}. However, in some situations, it is necessary that observations are assignable to more than one characteristic per dimension. For this reason, authors such as Wanner et al. (2021) \cite{Wanner2021}, Jöhnk et al. (2017) \cite{Johnk2017}, and Zschech (2018) \cite{Zschech2018} recommend omitting this restriction whenever reasonable. Therefore, we allow for the selection of multiple characteristics within one dimension to classify an observation if necessary. To develop the taxonomy, first, some meta-characteristics must be specified in advance and appropriate ending conditions must be selected. Meta-characteristics refer to the pivotal aspects derived from the taxonomy’s main purpose to ensure the actual relevance of the characteristics to be determined. According to Nickerson et al. (2013) \cite{Nickerson2013}, the previously mentioned ending conditions can be either subjective or objective but must be fulfilled compulsorily to stop the iterative development procedure and obtain the final taxonomy. Given these two preliminary necessities, the taxonomy development process can be initiated. Here, the authors confer the possibility of either choosing an empirical-to-conceptual or conceptual-to-empirical method. While the former starts with a subset of objects at hand to be classified and thus inductively derive the characteristics and dimensions, the latter is formed from prior knowledge about the domain of interest and thus ends with the assignment of the objects. After each iteration, the ending conditions are carefully evaluated regarding their fulfillment. If that is the case, the iterative taxonomy development process terminates at this stage and the derived taxonomy can be regarded as final. 

Moreover, we perform a cluster analysis to identify predominant patterns within the literature base regarding the taxonomy’s dimensions and characteristics to highlight typical application scenarios for synthetic data in advanced analytics and thereupon unveil possible avenues for future research. Cluster analyses are established confirmatory tools in academia to verify taxonomies \cite{Balijepally2011}. For this purpose, we assigned each article from the literature review to the characteristics of the taxonomy, that is, whenever a characteristic held true for an article, we assigned a “1”, else a “0”. Consequently, this procedure resulted in a vector representation of binary values (i.e., characteristics) for each article, which is highly suitable for clustering. Given the encoded literature, we perform a cluster analysis to inductively extract patterns in the data and thereby unveil typical application scenarios (e.g., \cite{Beinke2018,Napier2009,Ploesser2009}. In this context, agglomerative hierarchical clustering is a widely used approach to group objects based on their distance from each other \cite{Balijepally2011}. Here, a variety of possible clustering algorithms (e.g., single linkage, complete linkage, average linkage, centroid, or Ward’s method), as well as distance metrics (e.g., Euclidean, Yule, Hamming, or Dice), can be used \cite{Balijepally2011,Maronna2016}. Through exploration and careful evaluation of the resulting clusters, a decision must be taken at this stage to select an option and then elaborate on the identified types of application scenarios.

\section{Results}

\subsection{Data Collection}
\begin{figure}[H]
  \centering
  \includegraphics[width=0.8\textwidth]{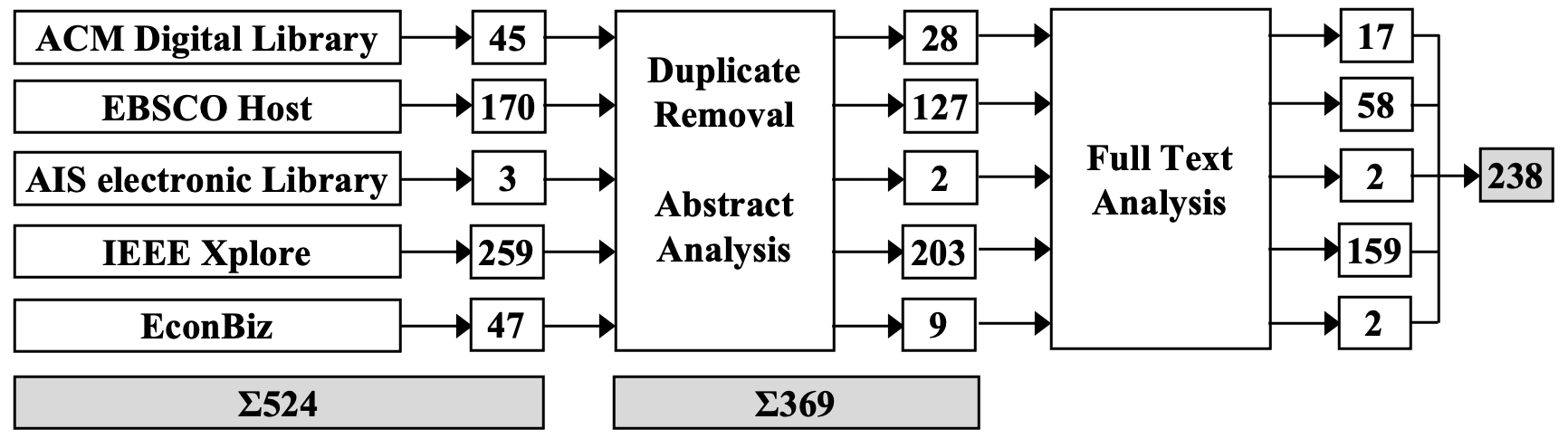}
  \caption{Overview of Literature Analysis.}
  \label{fig:fig4}
\end{figure}

As proposed by Cooper (1988) \cite{Cooper1988}, we first set an adequate review scope by using the suggested taxonomy to retrieve a literature base. Regarding the use of synthetic data in advanced analytics, questions arise as to why, how, and where such data are already used. In this context, three pillars are of special interest: (i) the motivation to use synthetic data, (ii) the generation method deployed, and (iii) the application field. Thus, we direct our focus towards (i) research methods and (ii) applications that generate and consume synthetic data. To this end, we do not concentrate on literature that is solely motivated by algorithmic challenges (e.g., improving image classification for the famous MNIST dataset by a decimal place) rather than finding a solution for a specific application context. Furthermore, we aim to identify and conceptually highlight central issues by taking a neutral perspective as described by Cooper (1988). The literature analysis is mainly targeted to a scientific audience, that is, scholars considering contributing to this specific field of research. Additionally, the review may provide value to the research community as a whole to gain a first broad conception of synthetic data as a topic to decide upon further investigation as well as to practitioners who are in charge of exploring new technologies or specifically concerned with the application of synthetic data in organizations. The literature is covered in a representative manner due to the choice of a certain search query and specific databases. To conceptualize the topic, we mark out the key elements based on our scope―i.e., (i) the motivation to use synthetic data, (ii) the generation method deployed, and (iii) the application field. These provide the contents sought within the literature review. Next, we scan five databases―namely, AIS electronic Library, IEEE Xplore, ACM Digital Library, EBSCO Host, EconBiz―for one of the following keywords in the articles’ titles to ensure a strong affiliation with the subject matter: “synthetic data” $\vee$ “synthesized data”. This yields a total of 524 articles as of February 28th 2022, of which 238 remain after full-text analysis (cf. Figure \ref{fig:fig4}). 

Besides the rather methodical cause (i.e., duplicates between the databases) this reduction is mainly the result of content-related considerations. For example, a large group of the studies relies on pre-existing synthetic data rather than creating their own proprietary dataset \cite{Chen2021}, whereas others do not provide any details on the data generation procedure \cite{Bue2010}. This phenomenon highlights the need for methodological guidance on the use of synthetic data. Moreover, as we are particularly interested in the application of synthetic data in different advanced analytics contexts, we discard publications that are solely concerned with the development or advancement of algorithms using established benchmark datasets (e.g., MNIST or CIFAR-10) without pointing towards a practical use case. This rationale leads to the exclusion of 73 articles from further analysis.

\begin{figure}[H]
  \centering
  \includegraphics[width=0.8\textwidth]{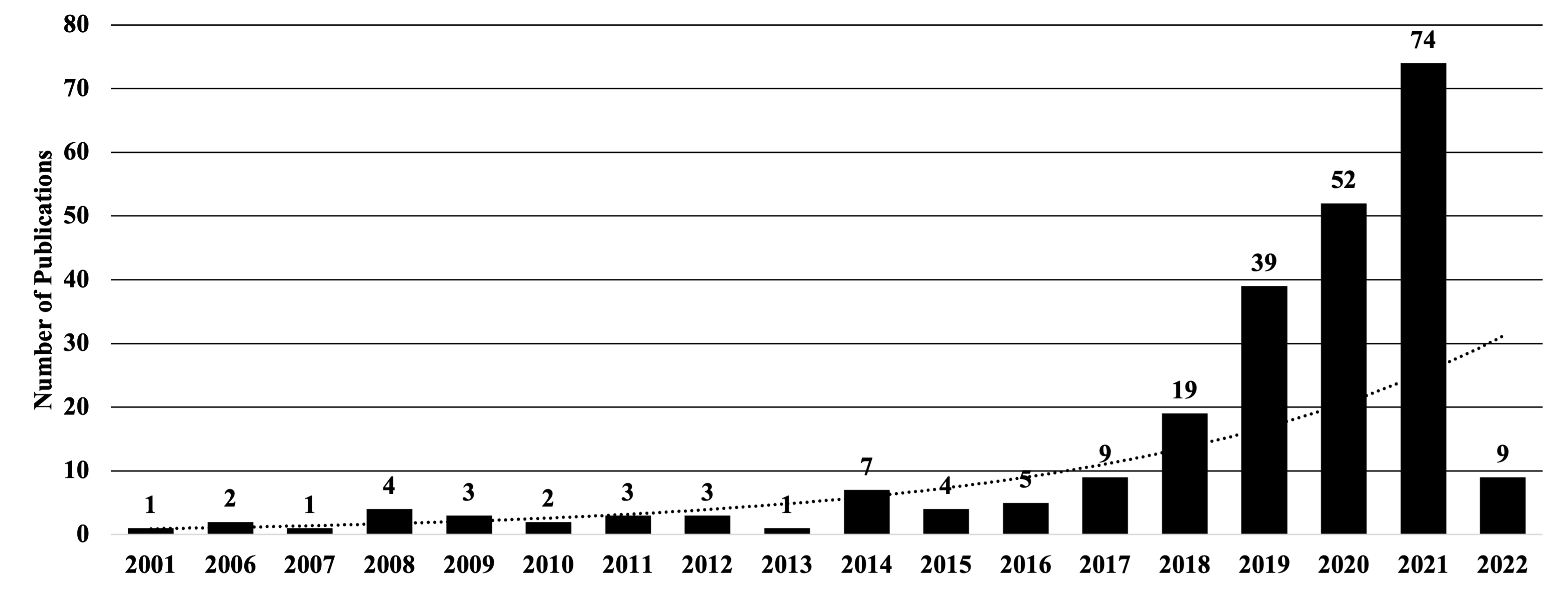}
  \caption{Number of Publications over Time (n=238).}
  \label{fig:fig5}
\end{figure}

Figure \ref{fig:fig5} depicts the number of relevant articles per year. The steeply growing number of publications―especially within the last three years―supports our initial assumption regarding the emergence of synthetic data as a topic in academia. 

\subsection{Taxonomy Development}
To derive the taxonomy, we first determine the meta-characteristics. Here, we rely on the focus of the previous literature review (i.e., the \textit{why}, the \textit{how}, and the \textit{where}). Accordingly, the selected meta-characteristics reflect the \textit{motivation} to make use of synthetic data, the \textit{generation} procedure, and the respective \textit{application}. Each of the dimensions and characteristics must apply to one of these meta-characteristics. Next, appropriate ending conditions, both subjective and objective, must be specified in advance to stop the iterative taxonomy development process (cf. Table \ref{tab:tab1}). 

\begin{table}[H]
 \caption{Overview on Fulfillment of Ending Conditions per Iteration}
  \centering
\begin{tabular}{cccccl}
\toprule
\multicolumn{5}{c}{\textbf{Iteration}}                         & \textbf{Ending condition}                                                    \\
\textbf{1} & \textbf{2} & \textbf{3} & \textbf{4} & \textbf{5} & \textbf{Objective condition}                                                 \\
\midrule

           & x          & x          & x          & x          & All relevant objects have been examined                                      \\
           & x          & x          & x          & x          & No further merge or split of objects                                         \\
           &            &            &            & x          & Each characteristic of each dimension was selected by at least one object    \\
           &            &            &            & x          & No new dimension or characteristic was added                                 \\
           &            &            &            & x          & No dimension was merged or split                                             \\
x          & x          & x          & x          & x          & Every dimension is unique                                                    \\
x          & x          & x          & x          & x          & Every characteristic is unique within its dimension                          \\
x          & x          & x          & x          & x          & Each cell is unique                                                          \\
\midrule
\multicolumn{5}{c}{}                                           & \textbf{Subjective condition}                                                \\
           &            & x          & x          & x          & Concise: Meaningful without being unwieldy overwhelming                      \\
           &            &            & x          & x          & Robust: The dimensions and characteristics enable differentiation of objects \\
           &            &            &            & x          & Comprehensive: All objects can be classified                                 \\
x          & x          & x          & x          & x          & Extensible: A new dimension or characteristic can be easily added            \\
x          & x          & x          & x          & x          & Explanatory: The dimensions and characteristics explain an object           \\
\bottomrule
\end{tabular}
\label{tab:tab1}
\end{table}

We then proceed to devise the taxonomy within five iterations. In the first iteration, we conduct the conceptual-to-empirical approach by building on the background of the various \textit{approaches} to generate synthetic data as our characteristics (i.e., \textit{statistical distributions}, \textit{simulation models}, \textit{data augmentation}, and \textit{DL-based}). As our ending conditions are not fulfilled at this point, we continue with a second iteration by using the empirical-to-conceptual approach. Here, we rely on extant research from the above survey and incorporate six \textit{expected benefits} from the use of synthetic data (i.e., \textit{extend dataset}, \textit{mitigate privacy issues}, \textit{reduce costs}, \textit{enhance data quality}, \textit{reduce effort}, and \textit{create specific data}) as a dimension for the first meta-characteristic. In addition, we determine two further dimensions in the literature to be associated with the generation of synthetic data aside from the \textit{approaches} themselves—namely, \textit{requirements} and \textit{synthetic portion}. While the former relates to the necessity of prior domain knowledge or initial data to create samples, the latter refers to the \textit{synthetic portion} of the data produced (i.e., either \textit{partially} or \textit{overly} synthetic). Since we did not cover insights on the last meta-characteristic (i.e., the application), we perform another empirical-to-conceptual iteration. Here, we add two new dimensions—again based on the findings of the literature review. These are the \textit{data type} (i.e., \textit{image}, \textit{tabular}, \textit{text}, and \textit{spectrogram}) and the \textit{usage context} (i.e., \textit{agriculture}, \textit{engineering and robotics}, \textit{security}, \textit{public service}, \textit{commerce}, \textit{media and graphical}, \textit{health services}, and \textit{transportation}). Given the taxonomy’s dedication to advanced analytics, in particular, an application’s \textit{sophistication degree} due to its considered added value may also be of interest to the user of the taxonomy. Thus, we switch back to the conceptual-to-empirical approach for the following iteration by incorporating the four levels (i.e., \textit{descriptive}, \textit{diagnostic}, \textit{predictive}, and \textit{prescriptive}) into the taxonomy as described by Banerjee et al. (2013) \cite{Banerjee2013} as well as Delen and Zolbanin (2018) \cite{Delen2018}. To verify the eligibility of these sophistication degrees, we once again opt for an empirical-to-conceptual iteration. We find that all four types of studies are represented in our literature base. Hence, we incorporate these four characteristics. By this point, we did not make any further changes to the dimensions, examined all the relevant objects, and met all other ending conditions (cf. Table \ref{tab:tab1}). Therefore, we can stop the iterative process at this stage.

The final taxonomy (cf. Table \ref{tab:tab2}) encompasses three meta-characteristics, seven dimensions, and 28 characteristics in total. As several expected benefits from the use of synthetic data may be present simultaneously, this dimension does not consist of mutually exclusive characteristics. Similarly, multiple approaches to the generation of synthetic data can be combined (e.g., a simulator may be used to render different scenes to be later augmented with image filters and then fed into a GAN for training an artificial synthetic image generator). Regarding the dimensions \textit{requirements} and \textit{synthetic portion} as well as the \textit{application} meta-characteristic, we in turn found every characteristic for each dimension to be mutually exclusive.

\begin{table}[h]
\caption{Taxonomy on the Use of Synthetic Data in Advanced Analytics}
  \centering
\resizebox{\linewidth}{!}{

\begin{tabular}{c|c|cccccccccc}
\textbf{Meta-characteristic} &
  \textbf{Dimension} &
  \multicolumn{10}{c}{\textbf{Characteristic}} \\ \hline
\textbf{Motivation} &
  Expected benefit &
  \multicolumn{1}{c|}{Extend dataset} &
  \multicolumn{3}{c|}{Mitigate privacy issues} &
  \multicolumn{1}{c|}{Reduce costs} &
  \multicolumn{1}{c|}{Enhance data quality} &
  \multicolumn{3}{c|}{Reduce effort} &
  Create specific data \\ \hline
\multirow{3}{*}{\textbf{Generation}} &
  Requirements &
  \multicolumn{5}{c|}{Domain knowledge} &
  \multicolumn{5}{c}{Initial data} \\ \cline{2-12} 
 &
  Approach &
  \multicolumn{3}{c|}{Statistical distribution} &
  \multicolumn{2}{c|}{Simulation model} &
  \multicolumn{2}{c|}{Data augmentation} &
  \multicolumn{3}{c}{DL-based} \\ \cline{2-12} 
 &
  Synthetic portion &
  \multicolumn{5}{c|}{Partially} &
  \multicolumn{5}{c}{Overly} \\ \hline
\multirow{3}{*}{\textbf{Application}} &
  Data type &
  \multicolumn{3}{c|}{Image} &
  \multicolumn{2}{c|}{Tabular} &
  \multicolumn{2}{c|}{Text} &
  \multicolumn{3}{c}{Spectrogram} \\ \cline{2-12} 
 &
  Usage context &
  \multicolumn{1}{c|}{Agriculture} &
  \multicolumn{2}{c|}{Commerce} &
  \multicolumn{1}{c|}{Security} &
  \multicolumn{1}{c|}{Public services} &
  \multicolumn{1}{c|}{Engineering and robotics} &
  \multicolumn{1}{c|}{Media and graphical} &
  \multicolumn{2}{c|}{Health services} &
  Transportation \\ \cline{2-12} 
 &
  Sophistication degree &
  \multicolumn{3}{c|}{Descriptive} &
  \multicolumn{2}{c|}{Diagnostic} &
  \multicolumn{2}{c|}{Predictive} &
   \multicolumn{1}{c}{} &
  \multicolumn{2}{c}{Prescriptive} \\ \hline
\end{tabular}}
\label{tab:tab2}

\end{table}

\subsection{Cluster Analysis}
\begin{figure}[h]
  \centering
  \includegraphics[width=0.99\textwidth]{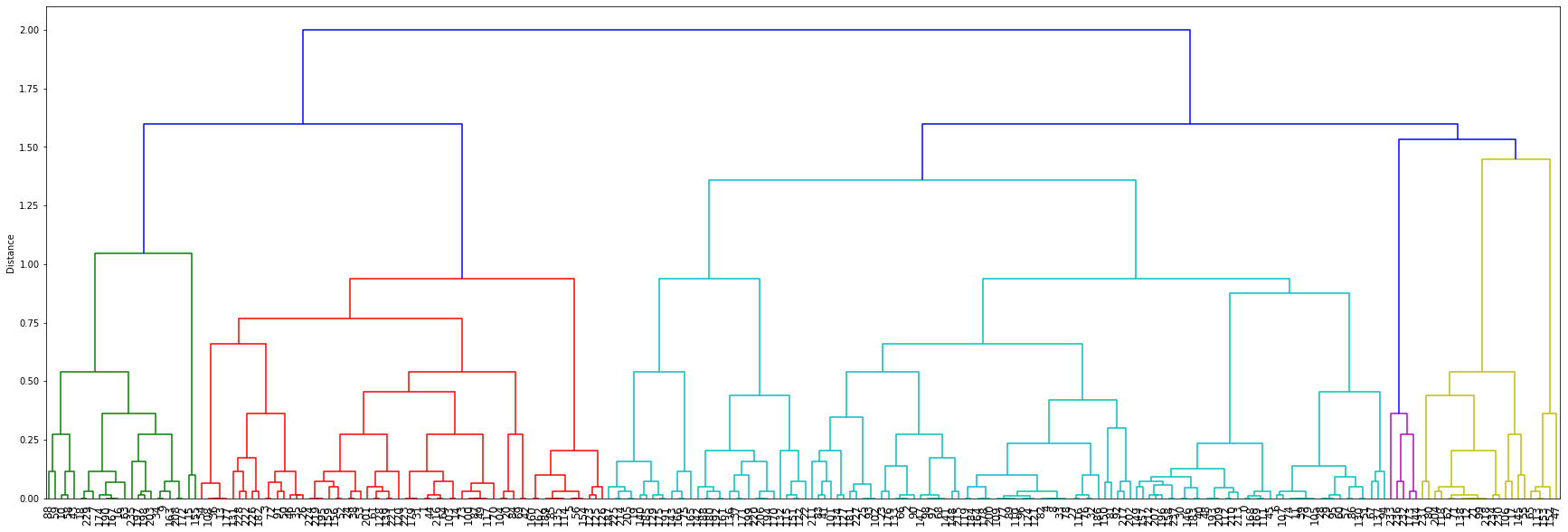}
  \caption{Dendrogram for Agglomerative Hierarchical Clustering.}
  \label{fig:fig6}
\end{figure}

Next, given the literature base and the final taxonomy, we proceed with the cluster analysis. To this end, we extensively explore and carefully evaluate the outputs of the various combinations of algorithms and metrics by intensively discussing them among five proficient researchers. We unanimously agreed on presenting the clusters that result from using agglomerative hierarchical clustering with the complete linkage algorithm and the Yule distance metric, as these are most meaningful and comprehensible. The corresponding dendrogram is depicted in Figure \ref{fig:fig6}, where each color represents one cluster.  

In addition, Table \ref{tab:tab3} provides detailed insights into the focal point of each identified cluster by summarizing the impact of each characteristic per subtheme in absolute and percentage values rounded to two decimal places. Next, we elaborate on these five clusters. 

Regarding the salient features of the clusters, we observe a significant difference in terms of the number of articles clustered together. More specifically, the smallest cluster comprises five studies whereas the third and largest identified subtheme includes more than half of the articles analyzed (i.e., 51.68\%, n=123). 

Regarding the peculiarities of the dimensions, the distribution of the characteristics represents the predominant uses of synthetic data in the literature. For instance, dataset enlargement is a frequently given reason (n=110) to employ synthetic data. Other popular expected benefits from the use of synthetic data are data quality improvements (n=80) and effort reduction (n=80). The less frequent motivators for the inclusion of synthetic data are cost reduction (n=46), privacy issue mitigation (n=42), and the creation of specific data required for the analysis (n=37). Notably, as for the generation approach, the characteristics’ distributions again reveal a tendency towards the use of simulation modeling (n=127) or data augmentation techniques (n=101) rather than the DL-based approaches (n=46) or sampling from statistical distributions (n=31). Whereas 150 out of 238 analyses rely on prior domain knowledge to create synthetic data, the remaining 88 studies leverage some initial data at their disposal. Remarkably, the number of studies using fully (n=121) or partially (n=111) synthetic datasets are quite comparable. The majority of the datasets underlying the studies’ analyses are image-based (n=148). With more than one-fourth (n=61) the studies concern tabular data while the remaining data types spectrographic (n=24) and textual (n=5) are rather underrepresented. The plethora of usage contexts of synthetic data is demonstrated via the respective distributions within the dimension. Health services (n=59) and the field of engineering and robotics (n=56) constitute the popular application areas, and public services (n=8) and commerce (n=15) are investigated less within the sample. Lastly, regarding the sophistication degree, most studies follow the predictive analytics paradigm (n=221), while the remaining articles pursue a prescriptive (n=11), descriptive (n=5), or diagnostic (n=1) endeavor. 

\textbf{Cluster 1 (C1) DL-based Synthetic Data Generation.} The first identified cluster comprises about one-tenth of the literature base (i.e., 10.08\%, n=24). Apart from one article, the cluster emphasizes DL-based data generation approaches using existing data only. Moreover, the applications developed here predominantly follow the predictive analytics paradigm (i.e., 91.67\%, n=22), and exactly two-thirds leverage tabular data as opposed to the other data types. The vast majority of the applications analyzed (i.e., 41.67\%, n=10) are related to health services. Notably, with 44 compared to 24, the amount of the expected benefits from using synthetic data exceeds the total objects in this cluster, which is due to the dimension not being mutually exclusive as discussed previously. Interestingly, none of the endeavors in C1 is motivated by creating specific synthetic data. Instead, the main motivation to deploy synthetic data is driven by extending pre-existing data (62.50\%), mitigating privacy-related issues (50\%), or enhancing the quality of the data at hand (45.83\%). For example, to overcome limited sensor data availability and quality (i.e., variety) in the context of healthcare applications, Dahmen and Cook (2019) \cite{Dahmen2019} employ hidden Markov models to produce adequate amounts of heterogenous synthetic data. The results attribute high effectiveness by using the additional data measured via the accuracy metric \cite{Dahmen2019}. Similarly, to encounter phishing attacks Shirazi et al. (2020) \cite{Shirazi2020} improve the detection of such malicious activities using additional data created with GANs. Likewise, their analysis amplifies that additional synthetic data can lead to higher predictive performance \cite{Shirazi2020}.

\textbf{Cluster 2 (C2) Image Data Augmentation for Predictive Analytics.} The next cluster (i.e., 26.90\%, n=64) is again almost completely represented by predictive analytics studies. However, as opposed to the previous subtheme, nearly four out of five articles deploy data augmentation techniques. Only every fourth article (i.e., 26.56\%, n=17) incorporates some of the DL-based approaches to synthetic data generation. Here the non-mutually exclusive criterion also applies such that the sum of the various approaches to synthetic data generation used (i.e., 71) exceeds the number of the articles clustered together. The other two approaches, namely, statistical distribution and simulation modeling, are barely deployed (i.e., 1 and 2, respectively). This leads to the conclusion that DL-based approaches are frequently deployed jointly with data augmentation techniques. As for the other dimensions requirements and synthetic portion, which are associated with the meta-characteristic generation, the numbers are on par, that is, prior domain knowledge is required for 3 out of 64 instances and likewise, overly synthetic data is created in 3 out of 64 articles. Another striking aspect of C2 is the predominant portion of studies considering the image data type (i.e., 81.81\%, n=53). A typical example of the articles clustered together in C2 is provided by Frid-Adar et al. (2018) \cite{Frid-Adar2018}. To overcome the limited amount of publicly available image data for liver lesion classification, the authors first apply classical data augmentation techniques and then train a GAN to later obtain entirely new images for the sake of improving the accuracy significantly \cite{Frid-Adar2018}. Similarly, Villalonga et al. (2020) \cite{Villalonga2020} propose the use of GANs to aid traffic sign recognition and classification to advance the field of autonomous vehicles.  

\begin{landscape}
\begin{table}[h]
 \caption{Result of Crosstab Analysis}
  \centering
\begin{tabular}{clccccccccccc}
    \toprule
\multicolumn{3}{c}{\textbf{}}                                                                     & \multicolumn{2}{c}{\textbf{Cluster 1}} & \multicolumn{2}{c}{\textbf{Cluster 2}} & \multicolumn{2}{c}{\textbf{Cluster 3}} & \multicolumn{2}{c}{\textbf{Cluster 4}} & \multicolumn{2}{c}{\textbf{Cluster 5}} \\
\textbf{Dimension}                     & \multicolumn{1}{c}{\textbf{Characteristic}} & \textbf{n} & \multicolumn{2}{c}{24}                 & \multicolumn{2}{c}{64}                 & \multicolumn{2}{c}{123}                & \multicolumn{2}{c}{5}                  & \multicolumn{2}{c}{22}                 \\
    \midrule
\multirow{6}{*}{Expected benefit}      & Enlarge dataset                             & 110        & 15              & 62.50\%              & 35              & 54.69\%              & 52               & 42.28\%             & 1              & 20.00\%               & 7               & 31.82\%              \\
                                       & Mitigate privacy issues                     & 42         & 12              & 50.00\%              & 7               & 10.94\%              & 8                & 6.50\%              & 0              & 0.00\%                & 15              & 68.18\%              \\
                                       & Reduce costs                                & 46         & 2               & 8.33\%               & 11              & 17.19\%              & 31               & 25.20\%             & 1              & 20.00\%               & 1               & 4.55\%               \\
                                       & Enhance data quality                        & 80         & 11              & 45.83\%              & 24              & 37.50\%              & 41               & 33.33\%             & 3              & 60.00\%               & 3               & 13.64\%              \\
                                       & Reduce effort                               & 88         & 4               & 16.67\%              & 31              & 48.44\%              & 51               & 41.46\%             & 0              & 0.00\%                & 3               & 13.64\%              \\
                                       & Create specific data                        & 37         & 0               & 0.00\%               & 5               & 7.81\%               & 32               & 26.02\%             & 2              & 40.00\%               & 1               & 4.55\%               \\
                                           \midrule

\multirow{2}{*}{Requirements}          & Domain knowledge                            & 150        & 0               & 0.00\%               & 3               & 4.69\%               & 121              & 98.37\%             & 4              & 80.00\%               & 22              & 100.00\%             \\
                                       & Initial data                                & 88         & 24              & 100.00\%             & 61              & 95.31\%              & 2                & 1.63\%              & 1              & 20.00\%               & 0               & 0.00\%               \\
                                           \midrule

\multirow{4}{*}{Approach}              & Statistical distribution                    & 31         & 1               & 4.17\%               & 1               & 1.56\%               & 11               & 8.94\%              & 0              & 0.00\%                & 17              & 77.27\%              \\
                                       & Simulation model                            & 127        & 0               & 0.00\%               & 2               & 3.13\%               & 114              & 92.68\%             & 5              & 100.00\%              & 7               & 31.82\%              \\
                                       & Data augmentation                           & 101        & 0               & 0.00\%               & 51              & 79.69\%              & 50               & 40.65\%             & 0              & 0.00\%                & 0               & 0.00\%               \\
                                       & DL-based                                    & 46         & 23              & 95.83\%              & 17              & 26.56\%              & 5                & 4.07\%              & 0              & 0.00\%                & 0               & 0.00\%               \\
                                           \midrule

\multirow{2}{*}{Synthetic portion}     & Partially                                   & 110        & 21              & 87.50\%              & 61              & 95.31\%              & 26               & 21.14\%             & 0              & 0.00\%                & 1               & 4.55\%               \\
                                       & Overly                                      & 121        & 3               & 12.50\%              & 3               & 4.69\%               & 97               & 78.86\%             & 5              & 100.00\%              & 21              & 95.45\%              \\
                                           \midrule

\multirow{4}{*}{Data type}             & Image                                       & 148        & 4               & 16.67\%              & 53              & 82.81\%              & 91               & 73.98\%             & 0              & 0.00\%                & 0               & 0.00\%               \\
                                       & Tabular                                     & 61         & 16              & 66.67\%              & 2               & 3.13\%               & 19               & 15.45\%             & 5              & 100.00\%              & 19              & 86.36\%              \\
                                       & Text                                        & 5          & 1               & 4.17\%               & 1               & 1.56\%               & 3                & 2.44\%              & 0              & 0.00\%                & 0               & 0.00\%               \\
                                       & Spectrogram                                 & 24         & 3               & 12.50\%              & 8               & 12.50\%              & 10               & 8.13\%              & 0              & 0.00\%                & 3               & 13.64\%              \\
                                           \midrule

\multirow{8}{*}{Usage context}         & Agriculture                                 & 20         & 3               & 12.50\%              & 3               & 4.69\%               & 10               & 8.13\%              & 1              & 20.00\%               & 3               & 13.64\%              \\
                                       & Commerce                                    & 15         & 2               & 8.33\%               & 4               & 6.25\%               & 5                & 4.07\%              & 1              & 20.00\%               & 3               & 13.64\%              \\
                                       & Security                                    & 24         & 5               & 20.83\%              & 1               & 1.56\%               & 17               & 13.82\%             & 0              & 0.00\%                & 1               & 4.55\%               \\
                                       & Public service                              & 8          & 2               & 8.33\%               & 1               & 1.56\%               & 2                & 1.63\%              & 0              & 0.00\%                & 3               & 13.64\%              \\
                                       & Engineering and robotics                    & 56         & 1               & 4.17\%               & 17              & 26.56\%              & 34               & 27.64\%             & 2              & 40.00\%               & 2               & 9.09\%               \\
                                       & Media and graphical                         & 31         & 1               & 4.17\%               & 15              & 23.44\%              & 13               & 10.57\%             & 0              & 0.00\%                & 2               & 9.09\%               \\
                                       & Health services                             & 59         & 10              & 41.67\%              & 17              & 26.56\%              & 23               & 18.70\%             & 1              & 20.00\%               & 8               & 36.36\%              \\
                                       & Transportation                              & 25         & 0               & 0.00\%               & 6               & 9.38\%               & 19               & 15.45\%             & 0              & 0.00\%                & 0               & 0.00\%               \\
                                           \midrule

\multirow{4}{*}{Sophistication degree} & Descriptive                                 & 5          & 0               & 0.00\%               & 0               & 0.00\%               & 0                & 0.00\%              & 3              & 60.00\%               & 2               & 9.09\%               \\
                                       & Diagnostic                                  & 1          & 0               & 0.00\%               & 0               & 0.00\%               & 0                & 0.00\%              & 0              & 0.00\%                & 1               & 4.55\%               \\
                                       & Predictive                                  & 221        & 22              & 91.67\%              & 63              & 98.44\%              & 120              & 97.56\%             & 1              & 20.00\%               & 15              & 68.18\%              \\
                                       & Prescriptive                                & 11         & 2               & 8.33\%               & 1               & 1.56\%               & 3                & 2.44\%              & 1              & 20.00\%               & 4               & 18.18\%        \\     
\bottomrule

\end{tabular}
  \label{tab:tab3}

\end{table}
\end{landscape}

\textbf{Cluster 3 (C3) Domain Knowledge for Data Simulation.} The third and largest identified subtheme includes more than half of the articles analyzed (i.e., 51.68\%, n=123). With 92.68\% (n=114), most of the endeavors rely on some form of simulation model, whereas data augmentation techniques account for 40.65\% (n=50) of the literature. This again indicates the non-mutually exclusiveness between the various data generation approaches and thus suggests possible value through their combined application. Shermeyer et al. (2021) \cite{Shermeyer2021}, for example, create a fully synthetic dataset for airplane classification through overhead images by using AI.Reverie’s generation platform. Here, the knowledge about weather situations, daytime, sunlight intensity, vantage point, biomes, and much more which is cast into the unreal engine is used by the image generation platform to model the bird's eye view aircraft images accordingly \cite{Shermeyer2021}. In addition, the authors apply blurring and cropping to improve data variability and thus robustness \cite{Shermeyer2021}. As for the other two approaches, statistical distribution (i.e., 8.94\%, n=11) and DL-based (i.e., 4.07\%, n=5) modeling, their importance for the studies grouped in C3 can be described comparably neglectable. At the outset, the researchers start with domain knowledge to generate synthetic data in 121 out of all 123 cases. Again, this subtheme predominantly comprises of predictive analytics endeavors (i.e., 97.56\%, n=120) considering image-based data (i.e., 73.98\%, n=91). The main drivers behind the creation of artificial data in C3 are the expected benefits from extending the dataset and the reduction of effort associated with an otherwise time-consuming and tedious data collection process. 

\textbf{Cluster 4 (C4) Overly Synthetic Tabular Data Simulation.} The fourth cluster represents the smallest (2.10\%, n=5). All the research endeavors grouped to C4 exclusively leverage simulation modeling to generate fully synthetic tabular data. Regarding the sophistication degree, however, three out of the five studies aim toward a descriptive analysis. Daud et al. (2012) \cite{Daud2012} simulate tabular data for seabed logging to descriptively provide insights into parameter variations for sediment thickness. To overcome data sparsity for the case of understanding navigation patterns on Wikipedia, Arora et al. (2022) \cite{Arora2022} construct synthetic navigation sequences via simulation modeling to obtain plausible clickstream data. 

\textbf{Cluster 5 (C5) Privacy-Preservation through Domain Knowledge.} The last identified subtheme accounts for 9.24\% (n=22) of the total sample. Here, the main benefit is expected from the mitigation of privacy issues (68.18\%, n=15). This is either done via statistical distribution (77.27\%, n=17) or simulation modeling (31.82\%, n=7). Interestingly, to generate the data, each study in C5 builds upon domain knowledge instead of initial data. For the most part, the cluster comprises studies that deal with tabular data (86.36, n=19) for health services. A typical example is given by Walonoski et al. (2020) \cite{Walonoski2020} . In their work, the authors rely on statistical distributions to simulate electronic health records with reference to the coronavirus for 124,150 patients and thereby evade privacy restrictions existing in the medical field \cite{Walonoski2020}.  

The results from the cluster analysis illustrate both the adoption of synthetic data in specific contexts (e.g., engineering and robotics) that has already taken place as well as the missed opportunities in that regard in the rather underrepresented areas (e.g., public service, security, agriculture, commerce, as well as media and graphical). It seems striking that most of the clusters, apart from C1, barely employ DL-based approaches to generating synthetic data. We attribute this lack in exploiting the promising opportunities to (i) the sheer novelty of these synthetic data generation methods as well as (ii) the lack of knowledge on the availability, functionality, or value of these methods, or even to (iii) the reluctancy on the part of the decision-makers. Surprisingly, for the usage contexts in C3, simulation models and augmentation operations are mostly employed in combination. As this cluster mainly comprises images, this is on par with the commonly observed proceed to first render some images in a simulation model to later augment them for various scenarios and, for example, enhance the robustness of the analytics application under development as demonstrated by Shermeyer et al. (2021) \cite{Shermeyer2021}. Although this approach may be sufficient for a variety of usage contexts, other analytics applications could take major advantage of the use of the more refined DL-based approaches due to the increased diversity of the synthetic data created (cf. \cite{Zhang2021,Esmaili2021,Baul2021,LeMinh2021,CAI2020}). Besides, to mitigate privacy restrictions for tabular data in healthcare applications, statistical distribution or simulation modeling are established options (cf. C5). Nevertheless, the exploration of DL-based approaches for this purpose might also be promising (e.g., \cite{Rankin2020}).

\section{Discussion}
The cluster analysis in particular reveals the variety of possible applications for synthetic data. Whether synthetic data is used to overcome a lack or shortage of information, privacy restrictions, quality issues, or time or cost constraints it holds the potential to drive analytics further. Against this backdrop, we argue that researchers and practitioners should be aware of the rich potential of synthetic data for all sorts of advanced analytics applications. Hence, they should deal with the expected benefits of using synthetic data by means of the devised taxonomy to discuss and carefully evaluate the usefulness for the considered application scenario in terms of effectiveness and efficiency. However, instead of exclusively attributing advantages to the use of synthetic data, we note that such data might induce uncertainty and therefore raise questions regarding the reliability of the results. Thus, to ensure adequate data quality, we recommend conducting continuous checks by employing appropriate metrics (e.g., \cite{Jordon2018}) or procedures (e.g., \cite{Chen2019a}). 

The generation itself is initiated either via domain knowledge or prior data already at disposal. Although the taxonomy provides a rather basic orientation with four distinct sets of generation approaches as opposed to a detailed breakdown of these, the cluster analysis already indicates the combined applicability of the methods. For example, oftentimes, a simulation model is utilized to create synthetic images, which afterward are transformed via augmentation techniques and sometimes additionally fed to DL-based approaches to increase the variety even further. However, we note that only a fraction of the studies analyzed (i.e., 18.91\%, n=45) with 36 in the last two years employ novel DL-based methods to produce synthetic data. This leads up to the conclusion, that the potentials associated are yet to be explored further to be fully unleashed. The created data can be either overly or partially synthetic as desired. A fully synthetic dataset may be suitable if privacy regulations are an issue, whereas a partially synthetic dataset may be appropriate if the advanced analytics application should be better in terms of performance or robustness. 

It is striking that most articles comprise predictive analytics studies as opposed to the other sophistication degrees. This, in turn, points toward more research required in that regard. For instance, descriptive statistics based on synthetic data may back communication with stakeholders in the prototyping phase of a project. Besides, synthetic data may help to answer questions on why an instance occurred and support investigating the underlying effects in the sense of diagnostic studies. Lastly, to determine an optimal action per situation, synthetic data acts as the enabler by providing the data necessary for the analysis.

\section{Conclusion}
The objective of the present research was to shed light on synthetic data as a promising topic that is increasingly gaining momentum in research (cf. data collection) due to major advances in data analytics as well as data generation driven by DL. Against the backdrop of the initially outlined limitations to using advanced analytics (i.e., limited or unavailable data) and the rather unattractive solutions for organizations to overcome these barriers (i.e., spending a substantial amount of money to gather more data; accepting the limited validity of the results; entirely refrain from advanced analytics activities), we argue that synthetic data holds great potential for advanced analytics and thus could become a key enabling factor for many of its applications. This may especially be the case for smaller organizations that are inevitably at a disadvantage compared to larger ones in terms of access to rich data sources. 

To encourage and support both researchers as well as practitioners to use synthetic data, we elaborated on a theoretically grounded and empirically validated, multidimensional taxonomy that provides guidance to answer the questions of \textit{why}, \textit{how}, and \textit{where} synthetic data can be applied. From a practical perspective, the developed taxonomy helps to systematically structure and communicate an endeavor. From a theoretical viewpoint, the taxonomy represents a pivotal point by outlining important aspects associated with synthetic data in a condensed form. Thus, researchers can use the taxonomy to study and hypothesize about relationships between the characteristics. Practitioners may apply the taxonomy to their usage context to carve out options for further development and discuss them in a structured manner with stakeholders.

The cluster analysis reveals popular options to benefit from the use of synthetic data in the context of advanced analytics. We empirically highlight typical application scenarios and plausible combinations of the taxonomy’s characteristics in the literature and, at the same time, uncover the rather unexplored options. This helps academics and practitioners estimate the potential and novelty of a specific endeavor. However, the cluster interpretation is not without its limitations either. The obtained results are heavily dependent on the chosen clustering algorithm and metric. To this end, we scrutinized our results qualitatively to ensure coherence and applicability. In the consecutive discussion, we reason about the opportunities to employ synthetic data for a variety of distinct usage contexts as well as the comparatively few studies that rely on DL-based data generation as opposed to the rather traditional approaches. 

Regarding the limitations of this article, first and foremost, the topic under investigation is evidently prone to fundamental changes due to rapid innovation in the field. Thus, with new possibilities (i.e., ground-breaking or more refined algorithms), new means to effectively enhance use cases through synthetic data may emerge. In addition, the results of the literature review are restricted by the database selection to be searched in and the use of a specific search query limited to the articles’ titles only. Furthermore, taxonomies serve as a starting point for contextualization and are therefore not complete per se. Although the taxonomy development process is rigorous and well-grounded, it does, by its nature, not necessarily account for all applications in the realm of synthetic data. Hence, with new generation capabilities and application domains emerging the taxonomy might be worth updating. Lastly, the cluster interpretation is not without its limitations either. The obtained results are heavily dependent on the chosen clustering algorithm and distance metric \cite{Balijepally2011}. To this end, we consistently scrutinized our results qualitatively to ensure coherence and applicability. 

Nevertheless, the present paper may open avenues for further research regarding the use of synthetic data in advanced analytics. First, it outlines the rather unexplored fields of application for synthetic data which might be very appealing to discover (e.g., public service, security, agriculture, commerce, as well as media and graphical). Secondly, it emphasizes the DL-based approaches to synthetic data generation but only focuses on two quite prominent methods―GANs and VAEs. Thus, the remaining methods might be worth exploring with respect to their potential for the different usage contexts―especially since these approaches are hardly or not at all applied in the literature examined. Furthermore, the many specific extensions and various implementations to these algorithms, as mentioned by Chen (2021) \cite{Chen2021} for instance, could hamper evaluating the fit-for-useness for the respective application field. Future research might provide detailed guidance upon choosing the right algorithms in terms of efficiency, effectiveness, and reliability (e.g., in the form of a toolbox). Lastly, as regards conducting advanced analytics projects, there is no common methodological approach that unifies the several opportunities on if, where (i.e., at which stage) and how (i.e., with which algorithm) to best incorporate synthetic data in dependence on the usage context.

\bibliographystyle{unsrt}  
\bibliography{references}

\end{document}